\documentclass[sigconf, nonacm]{acmart}
\AtBeginDocument{%
  \providecommand\BibTeX{{%
    \normalfont B\kern-0.5em{\scshape i\kern-0.25em b}\kern-0.8em\TeX}}}

\usepackage{subcaption}
\usepackage{amsthm}
\usepackage{graphicx}
\usepackage{multirow}
\usepackage{subcaption}

\theoremstyle{definition}
\newtheorem{definition}{Definition}

\newcommand{\Poincare}{Poincar\'e }
\newcommand{\Mobius}{M\"obius }
\newcommand{\algname}{HHGAT}

\begin{document}

\title{Hyperbolic Heterogeneous Graph Attention Networks}

%\title{Heterogeneous Graph Neural Networks with Enhanced Semantic Information Using Multi-Curvature Hyperbolic Space}

%\author{Anonymous Author(s)}
\author{Jongmin Park}
\affiliation{%
  \institution{Chungnam National University}
  \state{Daejeon}
  \country{South Korea}}
\email{pa5398@g.cnu.ac.kr}

\author{Seunghoon Han}
\affiliation{%
  \institution{Chungnam National University}
  \state{Daejeon}
  \country{South Korea}}
\email{tmdgns129@o.cnu.ac.kr}

\author{Soohwan Jeong}
\affiliation{%
  \institution{Chungnam National University}
  \state{Daejeon}
  \country{South Korea}}
\email{integerhwan@g.cnu.ac.kr}

\author{Sungsu Lim}
\authornote{Corresponding author.}
\affiliation{%
  \institution{Chungnam National University}
  \state{Daejeon}
  \country{South Korea}}
\email{sungsu@cnu.ac.kr}
\renewcommand{\shortauthors}{Jongmin Park, Seunghoon Han, Soohwan Jeong, and Sungsu Lim}
%\renewcommand{\shortauthors}{Anonymous Author(s)}

\begin{abstract} 
Most previous heterogeneous graph embedding models represent elements in a heterogeneous graph as vector representations in a low-dimensional Euclidean space. However, because heterogeneous graphs inherently possess complex structures, such as hierarchical or power-law structures, distortions can occur when representing them in Euclidean space. To overcome this limitation, we propose Hyperbolic Heterogeneous Graph Attention Networks (\algname{}) that learn vector representations in hyperbolic spaces with metapath instances. We conducted experiments on three real-world heterogeneous graph datasets, demonstrating that \algname{} outperforms state-of-the-art heterogeneous graph embedding models in node classification and clustering tasks. 
%This superior performance is attributed to \algname{}'s ability to capture the complex structure of heterogeneous graphs effectively.
\end{abstract}


\iffalse


\fi
\maketitle
\begin{figure}[]
\centering
\includegraphics[width=\columnwidth]{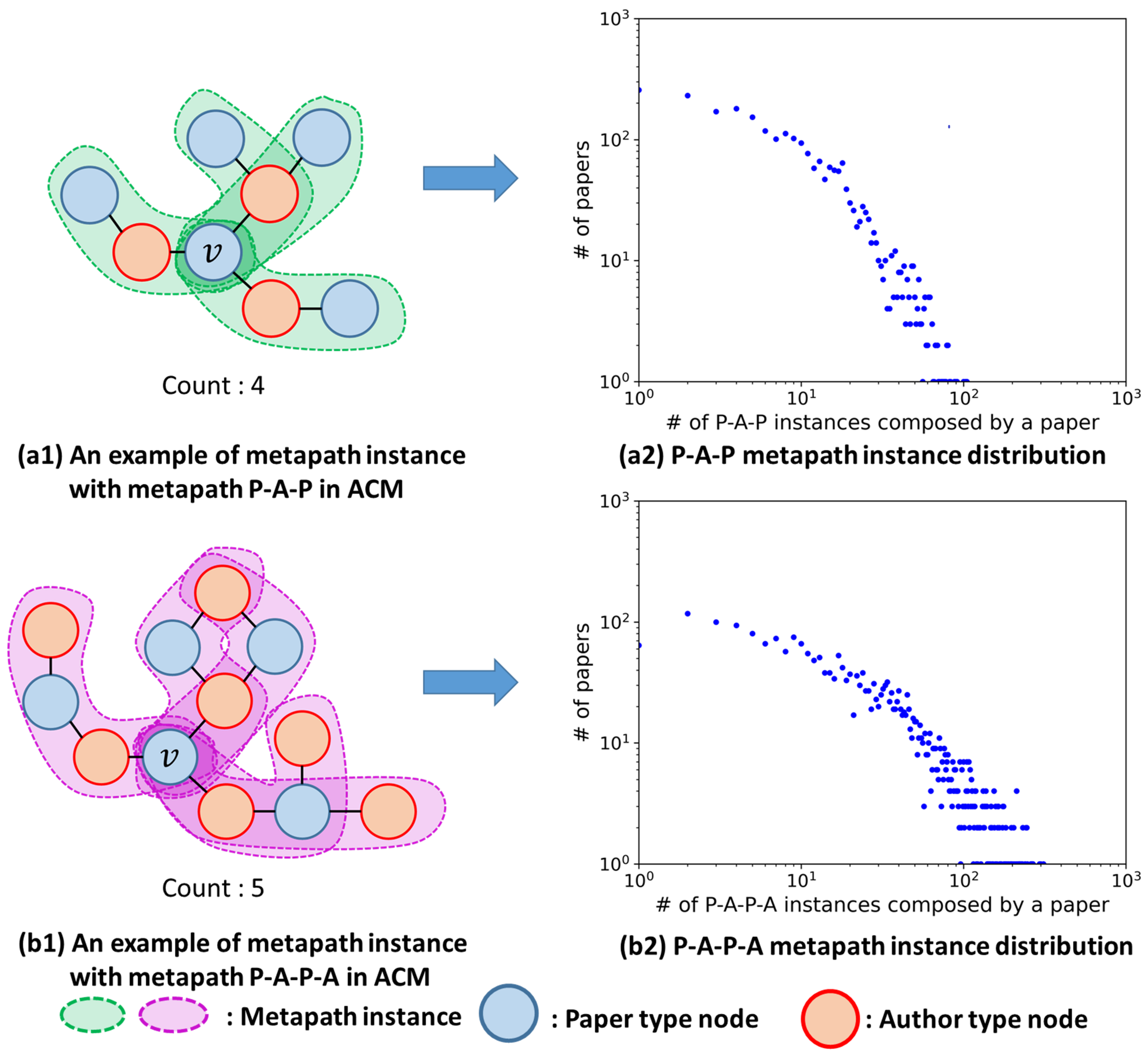}
\vspace{-1em}
\caption{Examples of metapath instances and metapath instance distribution in ACM dataset.}
\label{fig:Data_distribution}
\vspace{-2.5em}
\end{figure}

\section{Introduction}
Heterogeneous graphs consist of multiple types of nodes and links, enabling effective modeling of real-world scenarios. To address machine learning problems involving heterogeneous graphs, recent studies~\cite{wang2019www, fu2020www, hu2020, yun2019neurips, li2021aaai, park2023migtnet} aiming to represent elements in a heterogeneous graph as vector representations in a low-dimensional Euclidean space have adopted graph neural networks. HAN~\cite{wang2019www} leverages information from metapath-based neighbors. In HAN, the intermediate nodes in the metapath instance are ignored, whereas MAGNN~\cite{fu2020www} extends HAN by considering these intermediate nodes. 
GTN~\cite{yun2019neurips} learns a soft selection by stacking graph transformer layers to learn metapaths.
%HGT~\cite{hu2020} proposes node and edge type-dependent attention mechanisms to capture the various types of relations within the heterogeneous graph. 
%Simple-HGN~\cite{lv2021we} extends graph attention mechanisms by calculating attention scores through incorporating edge type information.

While these studies achieve notable performance, is Euclidean space truly appropriate for heterogeneous graphs? Recent studies~\cite{wang2019hyperbolic, wang2021embedding, li2023multi} on heterogeneous graph embedding show that real-world heterogeneous graphs have complex structures, including hierarchical and power-law structures. Therefore, they argue that hyperbolic space, which naturally inherits properties of complex structures, is more appropriate for heterogeneous graph embedding than Euclidean space. HHNE~\cite{wang2019hyperbolic} proposed a metapath-based random walk strategy for sampling metapath instances and embedding them into hyperbolic space, demonstrating the effectiveness of learning metapath instances in hyperbolic spaces. SHAN~\cite{li2023multi} proposed heterogeneous graph neural networks with hyperbolic space to learn representations of heterogeneous graphs, leveraging simplicial complexes and hyperbolic graph attention mechanisms for multi-order relations. However, the limitations of these studies are that HHNE is a shallow embedding model that cannot fully leverage node features. On the other hand, SHAN extracts a simplicial complex consisting of nodes of the same type to learn multi-order relations, it does not effectively capture the heterogeneity inherent in relations between different types of nodes explicitly represented in heterogeneous graphs.

In a heterogeneous graph, a metapath is defined as an ordered sequence of node or link types, and it reflects the semantic information of a heterogeneous graph. In addition, a metapath instance is defined as a node sequence in a heterogeneous graph following the schema defined by the metapath. From metapath instances, we can capture not only rich semantic structural information within a heterogeneous graph but also heterogeneity from various types of nodes within metapath instances. Figure~\ref{fig:Data_distribution} illustrates examples of metapath instances starting from node $v$ and demonstrates that these metapath instances follow a power-law distribution in real-world heterogeneous graphs. For example, P-A-P metapath instances reveal when two papers share an author, while P-A-P-A instances additionally indicate co-authorship. Hence, we can observe heterogeneity between different types of nodes within each metapath instance.

To overcome the limitations of previous studies, we propose the Hyperbolic Heterogeneous Graph Attention Network (\algname{}), designed to learn rich semantic structural information from metapath instances following power-law distributions.
Specifically, \algname{} automatically samples metapath instances within the maximum metapath length based on given link types to address the challenge of defining metapaths relying on domain-specific knowledge and leverages them in hyperbolic spaces to learn heterogeneous graph representations. 

%By doing so, \algname{} effectively represents the heterogeneous semantic structural information within different types of metapaths into hyperbolic space. Finally, \algname{} aggregates these representations to learn the heterogeneity of various semantic information within a heterogeneous graph.

%To overcome these limitations, \jong{we propose the Hyperbolic Heterogeneous Graph Attention Network (\algname{}), which is designed to learn rich semantic structural information within metapath instances following power-law distributions.} Figure~\ref{fig:Data_distribution} illustrates the distributions of metapath instances for certain metapaths in the ACM dataset, demonstrating a power-law distribution  Due to this property, inherent in heterogeneous graphs, \algname{} samples metapath instances within the maximum metapath length according to given link types and leverages them in the hyperbolic space. Note that \algname{} generates metapaths without domain knowledge. \jong{By leveraging metapath instances, \algname{} effectively represents the heterogeneous semantic structural information within different types of metapaths into hyperbolic space. Additionally, aggregating these representations allows us to learn the heterogeneity of various semantic information within a heterogeneous graph.}

%Figure~\ref{fig:Data_distribution} illustrates the distributions of metapath instances for some metapaths in the ACM dataset, demonstrating a power-law distribution.

Our contributions are summarized as follows:
\begin{itemize}
%\vspace{-0.4em}
    \item To the best of our knowledge, we are the first to propose hyperbolic heterogeneous graph neural networks for learning metapath instances. \algname{} can effectively learn the hierarchical structure of metapath instances explicitly present in heterogeneous graphs.
    \item We propose attention mechanisms in hyperbolic spaces to enhance the learning of node representations in heterogeneous graphs.
    \item We empirically show that \algname{} outperforms state-of-the-art algorithms in node classification and clustering tasks. In addition, we analyzed the effectiveness of the curvature parameter in hyperbolic spaces.
\end{itemize}
\vspace{-0.5em}
\section{Preliminaries}
\vspace{-0.3em}
\begin{definition}[\bf \Poincare ball model]
The \Poincare ball model with curvature $-c\ (c>0)$ is defined by the Riemannian manifold $(\mathbb{D}^{n,c}, g_x^c)$, where
\vspace{-0.5em}
\begin{align*}
     \mathbb{D}^{n,c}&=\{x \in \mathbb{R}^n : c||x||^2 < 1\},\;g_x^c = (\lambda^c_x)I_d.
\end{align*}
Here, $\mathbb{D}^{n,c}$ is the open $n$-dimensional ball with radius $\frac{1}{\sqrt{c}}$, $g_x^c$ is the Riemannian metric tensor where $\lambda_x^c=\frac{2}{1-c||x||^2}$, $I_d$ is the identity matrix. We denote $\mathcal{T}_x\mathbb{D}^{n,c}$ as the tangent space of $\mathbb{D}^{n,c}$ at $x$.
\end{definition}
\vspace{-0.3em}
\begin{definition}[\bf \Mobius addition] Given $x,y \in \mathbb{D}^{n,c}$, the \Mobius addition,
representing the addition operation in the \Poincare ball $\mathbb{D}^{n,c}$, is defined as follows:
%which represents the equation for the addition operation in the \Poincare ball model with curvature $-c$ is defined as follows:
\begin{align*}
    x\oplus_c y &= \frac{(1+2c\langle x,y\rangle+c||y||^2)x+(1-c||x||^2)y}{1+2c\langle x,y \rangle+c^2||x||^2||y||^2},
\end{align*}
where $\langle\cdot\rangle$ and $||\cdot||$ represent the Euclidean inner product and norm, respectively.
\end{definition}
\vspace{-0.3em}
\begin{definition}[\bf Exponential and logarithmic maps]
The exponential map $\text{exp}^c_x:\mathcal{T}_x\mathbb{D}^{n,c}\rightarrow\mathbb{D}^{n,c}$ and the logarithmic map $\text{log}^c_x:\mathbb{D}^{n,c}\rightarrow \mathcal{T}_x\mathbb{D}^{n,c}$ are defined as follows:
\vspace{-0.5em}
\begin{align*}
    \text{exp}_x^c(v) &= x\oplus_c\left(\text{tanh}\left(\sqrt{c}\frac{\lambda_x^c||v||}{2}\right)\frac{v}{\sqrt{c}||v||}\right),\\
    \text{log}_x^c(y)&=\frac{2}{\sqrt{c}\lambda_x^c}\text{tanh}^{-1}(\sqrt{c}||-x\oplus_cv||)\frac{-x\oplus_c y}{||-x\oplus_c y||},
\end{align*}
\end{definition}
\noindent where $x$ and $y$ are points in the hyperbolic space $\mathbb{D}^{n,c}$ and $x\neq y$. $v$ is a nonzero tangent vector in the tangent space $\mathcal{T}_x\mathbb{D}^{n,c}$.
\vspace{-0.3em}
\begin{definition}[\bf Hyperbolic matrix-vector multiplication] 
Given $x\in\mathbb{D}^{n,c}$ and a weight matrix $M\in\mathbb{R}^{m\times n}$, the matrix multiplication operation in hyperbolic spaces is defined as follows: if $Mx\neq 0$,
\begin{align*}
    M\otimes_c x &= \text{exp}_0^c(M\text{log}_0^c(x)).%(\mathbf{0}\in\mathbb{R}^{n}).
\end{align*} 
\end{definition}
\vspace{-0.3em}
\begin{definition}[\bf Hyperbolic non-linear activation function]
Given the point $x\in\mathbb{D}^{n,c}$, the hyperbolic non-linear activation function is defined as follows:
\vspace{-1em}
\begin{align*}
    \sigma\otimes^c(x) &= \text{exp}_0^c(\sigma(\text{log}_0^c(x))),
\end{align*}
where $\sigma$ is a Euclidean non-linear activation function.
\end{definition}

\section{Methodology}
\subsection{Metapath Instance Sampling}
Given a set of embedding target nodes $\mathcal{V}_t$, \algname{} samples the set of metapath instances consisting of metapath instances that start from node $v\in\mathcal{V}_t$ and have lengths within the maximum metapath length $l$. Each metapath instance $p\in\mathcal{M}_v$ follows metapath $\phi\in\Phi$, a set of metapaths. In this sampling process, we employ breadth-first search, which effectively captures the hierarchical structure around node $v$. Note that this procedure doesn't require predefined metapaths.

\subsection{Metapath-specific Embedding}
\subsubsection{\bf Euclidean Metapath Instance Feature}
Equation~\ref{eq:1} denotes the concatenation operation of the features of nodes $x_i\in\mathbb{R}^n$ within a metapath instance, where $j$ represents the length of metapath instance $p$.
\algname{} concatenates all $x_i$ within $p$ to preserve information of intermediate nodes and their order. We then consider this concatenated feature as the feature of the metapath instance.
\begin{align}
\label{eq:1}
    x_{p}^\mathbb{E} &= \; \parallel^j_{i=1}x_i\;(j\leq l).
\end{align}

\subsubsection{\bf Hyperbolic metapath instance embedding} For hyperbolic metapath instance embedding, \algname{} first maps Euclidean feature of metapath instance $x_{p}^\mathbb{E}$ to hyperbolic space $\mathbb{D}^{jd,c}$. We can assume that $x_{p}^\mathbb{E}$ is included in the tangent space $\mathcal{T}_x\mathbb{D}^{jd,c}$ at $x=0$, then we can map $x_{p}^\mathbb{E}$ to hyperbolic space via exponential map $\text{exp}_o^{c}$, where $o\coloneqq\mathbf{0}\in\mathbb{R}^{jd}$ that denotes origin manifold $\mathbb{D}^{jd,c}$. A hyperbolic feature of metapath instance $p$, denoted as $x_p^\mathbb{H} \in \mathbb{D}^{jd,c}$, is represented as follows:
\vspace{-1em}
\begin{align}
    x_p^\mathbb{H} &= \text{exp}_o^{c}(x_p^\mathbb{E}),
\end{align}
where $-c$ is a learnable parameter that denotes the negative curvature of hyperbolic space.

After mapping from Euclidean to hyperbolic space, we use hyperbolic linear transformation to map $x_p^\mathbb{H}$ into the embedding space. The formulation of a hyperbolic linear transformation is as follows:
\vspace{-1em}
\begin{align}
\label{eq:3}
    h_p^\mathbb{H} &= (W\otimes_{c}x_p^\mathbb{H})\oplus \text{exp}_o^{c}(b).
\end{align}

In Equation~\ref{eq:3}, $W\in\mathbb{R}^{d'\times jd}$ is the weight matrix corresponds to the metapath instances following the metapath $\phi$, $d'$ is the dimension of metapath-specific embedding vector and $h_p^\mathbb{H}\in\mathbb{D}^{d',c}$ is the hyperbolic metapath instance embedding vector.
\subsubsection{\bf Metapath Instance Aggregation} The attention mechanism in hyperbolic spaces is used to learn the attention score of each metapath instance and aggregate them accordingly. In this process, we calculate the attention score of each metapath instance $\alpha_p$ as follows:
\vspace{-1em}
\begin{align}
\label{eq:4}
    i_p &= \sigma\otimes^{c}(a^T\cdot(\text{log}_o^{c}(h_p^\mathbb{H})),\\
    \alpha_p &= \frac{\text{exp}(i_p)}{\sum_{q\in\mathcal{M}_v^\phi}(\text{exp}(i_q))}, 
\end{align}
where $a\in\mathbb{R}^{d'}$ is the attention vector.

Once the attention scores for each metapath instance following a specific metapath $\phi$ are obtained, \algname{} performs metapath instance aggregation to obtain the metapath-specific embedding vector $z_v^\phi\in\mathbb{D}^{d',c}$ as follows:
\vspace{-0.5em}
\begin{align}
\label{eq:6}
    z_v^\phi &= \sigma\otimes^{c}(\text{exp}_o^{c}(\sum_{p\in\mathcal{M}_v^\phi}(\alpha_p\cdot \text{log}_o^{c}(h_p^\mathbb{H})))).
\end{align}
In Equations~\ref{eq:4} and~\ref{eq:6}, $\sigma\otimes^{c}(\cdot)$ denotes the hyperbolic non-linear activation function with ReLU.

\begin{table}[t!]
\centering
\caption{Statistics of datasets}
\vspace{-1.5em}
\label{tab:datastats}
\resizebox{\columnwidth}{!}{%
\begin{tabular}{cccccc} \hline
Dataset & \# Nodes & \# Links & \# Class & Splitting  & \# Features \\\hline
IMDB    & \begin{tabular}[c]{@{}c@{}}Movie (M) : 4,661\\ Director (D) : 2,270\\ Actor (A) : 5,841\end{tabular}   & \begin{tabular}[c]{@{}c@{}}M-D : 4,661\\ M-A : 13,983\end{tabular} & 3        & \begin{tabular}[c]{@{}c@{}}\# Train : 1,410\\ \# Valid : 353\\ \# Test : 1,175\end{tabular} & 1,256\\\hline
DBLP    & \begin{tabular}[c]{@{}c@{}}Author (A) : 4,057\\Paper (P) : 14,328\\ Conference (C) : 20\end{tabular}   & \begin{tabular}[c]{@{}c@{}}A-P : 19,645\\ P-C : 14,328\end{tabular} & 4       & \begin{tabular}[c]{@{}c@{}}\# Train : 1,947\\ \# Valid : 487\\ \# Test : 1,622\end{tabular} & 334\\\hline
ACM    & \begin{tabular}[c]{@{}c@{}}Paper (P) : 3,020\\ Author (A) : 5,912\\ Subject (S) : 57\end{tabular}   & \begin{tabular}[c]{@{}c@{}}P-A : 9,936\\ P-S : 3,025\\\end{tabular} & 3        & \begin{tabular}[c]{@{}c@{}}\# Train : 1,452\\ \# Valid : 363\\ \# Test : 1,209\end{tabular} & 1,902\\\hline
\end{tabular}%
}
\vspace{-1em}
\end{table}
\vspace{-1em}
\subsection{Inter-metapath Attention}
\label{inter-metapath attention}
For given metapath-specific embedding vectors, \algname{} aggregates them using an attention mechanism. The attention score for each metapath-specific embedding vector $\beta_\phi$ can be calculated as follows:
\vspace{-1em}
\begin{align}
    \label{eq:7}
    w_\phi &= q^T\cdot\text{tanh}(M\cdot \text{log}_o^{c}(z_v^\phi)+b),\\
    \label{eq:8}
    \beta_\phi &= \frac{\text{exp}(w_\phi)}{\sum_{\pi\in\Phi}(\text{exp}(w_\pi))}.
\end{align}

In Equation~\ref{eq:7}, $M\in\mathbb{R}^{d_e\times d'}$ is the weight matrix corresponds to $z_v^\phi$, and $q\in\mathbb{R}^d_e$ is the attention vector and $d_e$ is the dimension of final node embedding vector.
With the learned attention score for metapath-specfic embedding vectors, \algname{} can assign higher weights to semantically important metapath-specific embedding vectors. The final node embedding for node $v$, denoted as $Z_v\in\mathbb{R}^{d_e}$, is represented in Equation~\ref{eq:9}. Here, $\sigma$ represents the ReLU activation function.
\begin{align}
\label{eq:9}
    z_v &= \sigma(\sum_{\phi\in\Phi}(\beta_\phi\cdot \text{log}_o^{c}(z_v^\phi)).
\end{align}
\vspace{-1em}
\subsection{Model Training}
We utilize the linear transformation $f(\cdot)$ to project node embedding vectors into a space with the specified output dimension. This transformation is formulated as follows:
\begin{align}
    f(z_v) &= \sigma(W_c\cdot z_v),
\end{align}
where $W_c \in \mathbb{R}^{d_o \times d_e}$ is the weight matrix with $d_o$ representing the dimension of the output vector and $\sigma$ is the activation function. 
Then, \algname{} is trained by minimizing the cross-entropy function $\mathcal{L}$, which is defined as follows:
\begin{align}
    \mathcal{L} &=-\sum_{v\in V_t}\sum_{c=1}^C(y_v[c]\cdot\text{log}(f(z_v)[c])),
\end{align}
where $v_t$ is the embedding target node set from labeled node set, $C$ is the number of classes, $y_v$ is the one-hot encoded label vector of node $v$, and $f(z_v)$ is a vector representing the label probabilities of node $v$.
\vspace{-1em}
\section{Experiments}
\subsection{Datasets}
We conducted experiments using three widely-used real-world heterogeneous graph datasets provided by GTN~\cite{yun2019neurips}. Table~\ref{tab:datastats} shows the statistics of the datasets. IMDB is an online database related to movies and television programs. Movie type of nodes were labeled according to the movie's genre. In DBLP and ACM, both of which are citation networks, the Author and Paper types of nodes are labeled according to the author's research area and the paper's subject area, respectively. 
\vspace{-0.75em}

\subsection{Baselines}
We compare \algname{} with several state-of-the-art GNN models. These baselines are divided into four categories: Euclidean homogeneous GNNs including GCN~\cite{kips2017iclr} and GAT~\cite{velickovic2018iclr}; hyperbolic homogeneous GNNs including HGCN~\cite{chami2019hyperbolic}; Euclidean heterogeneous GNNs including HAN~\cite{wang2019www}, MAGNN~\cite{fu2020www}, GTN~\cite{yun2019neurips}, HetSANN~\cite{hong2020attention}, HGT~\cite{hu2020}, GraphMSE~\cite{li2021aaai} and Simple-HGN~\cite{lv2021we}; hyperbolic heterogeneous GNNs including SHAN~\cite{li2023multi}. Note that, for a fair comparison between homogeneous GNNs and heterogeneous GNNs, we preprocess the node features to be homogeneous.
\vspace{-1em}

\subsection{Implementation Details}
For the baselines, we randomly initialize parameters and use the Adam optimizer with a learning rate of 0.0001 and a weight decay of 0.001. 
We set the dropout rate to 0.5, the embedding dimension to 64, and the number of attention heads to 8 for multi-head attention-based models. The baseline models are trained for 100 epochs, and we adopt early stopping with patience of 20 epochs. The metapath settings for metapath-based heterogeneous GNNs follow the specifications outlined in their papers. For \algname{}, the maximum length of metapath $l$ is set to 4, 5, and 3 for IMDB, DBLP, and ACM, respectively.

\begin{table*}[]
\centering
\caption{Experimental results (\%) for the node classification task.}
\vspace{-1.5em}
\label{tab:node_classification}
\resizebox{\textwidth}{!}{%
\begin{tabular}{|c|c|cccccccccccc|}
\hline
\multirow{2}{*}{Dataset} &
  \multirow{2}{*}{Metric} &
  \multicolumn{12}{c|}{Baselines} \\ \cline{3-14} 
 &
   &
  GCN &
  GAT &
  HGCN &
  HAN &
  MAGNN &
  GTN &
  HetSANN &
  HGT &
  GraphMSE &
  Simple-HGN &
  SHAN &
  \algname{} \\ \hline
\multirow{2}{*}{IMDB} &
  Macro-F1 &
  54.37$\pm$0.43 &
  56.91$\pm$0.28 &
  57.89$\pm$0.59 &
  58.92$\pm$0.51 &
  60.72$\pm$0.46&
  61.79$\pm$0.61&
  60.21$\pm$0.58 &
  60.63$\pm$0.20 &
  63.57$\pm$0.43 &
  64.72$\pm$0.75 &
  66.63$\pm$0.73 &
  \underline{\bf67.01$\pm$0.55} \\
 &
  Micro-F1 &
  54.53$\pm$0.41 &
  56.18$\pm$0.35&
  57.91$\pm$0.48 &
  58.44$\pm$0.86&
  60.62$\pm$0.63&
  63.63$\pm$0.71 &
  57.14$\pm$0.56&
  62.74$\pm$0.49&
  66.81$\pm$0.52&
  67.12$\pm$0.76&
  69.59$\pm$0.80 &
  \underline{\bf 70.23$\pm$0.54}
   \\ \hline
\multirow{2}{*}{DBLP} &
  Macro-F1 &
  89.42$\pm$0.23&
  91.50$\pm$0.48&
  92.58$\pm$1.06&
  92.83$\pm$0.51&
  93.62$\pm$0.64&
  93.75$\pm$0.42&
  92.15$\pm$0.54&
  92.27$\pm$0.49&
  94.35$\pm$0.31&
  94.02$\pm$0.43&
  94.53$\pm$0.25 &
  \underline{\bf 95.72$\pm$0.27}
   \\ 
 &
  Micro-F1 &
  89.16$\pm$0.51 &
  92.17$\pm$0.43 &
  93.36$\pm$0.87 &
  93.33$\pm$0.63 &
  94.13$\pm$0.51 &
  94.25$\pm$0.38 &
  92.78$\pm$0.56 &
  92.61$\pm$0.18 &
  94.50$\pm$0.56 &
  94.70$\pm$0.51 &
  94.67$\pm$0.26 &
  \underline{\bf 96.06$\pm$0.30}
   \\ \hline
\multirow{2}{*}{ACM} &
  Macro-F1 &
  88.51 $\pm$0.49&
  88.39$\pm$0.37 &
  89.95$\pm$0.42 &
  91.23$\pm$0.31 &
  90.69$\pm$0.29 &
  91.27$\pm$0.19 &
  89.20$\pm$0.27 &
  90.88$\pm$0.17 &
  93.34$\pm$0.30 &
  93.32$\pm$0.19 &
  93.90$\pm$0.31 &
  \underline{\bf 94.40$\pm$0.23}
   \\
 &
  Micro-F1 &
  88.22$\pm$0.45 &
  87.86$\pm$0.43 &
  90.24$\pm$0.37 &
  92.44$\pm$0.32 &
  92.15$\pm$0.25 &
  92.05$\pm$0.23 &
  90.02$\pm$0.35 &
  91.23$\pm$0.39 &
  93.31$\pm$0.27 &
  93.36$\pm$0.51 &
  94.42$\pm$0.30 &
  \underline{\bf 94.45$\pm$0.21}
   \\ \hline
\end{tabular}%
}
\vspace{-1em}
\end{table*}

% \usepackage{multirow}
% \usepackage{graphicx}
\begin{table*}[]
\centering
\caption{Experimental results (\%) for the node clustering task.}
\vspace{-1.5em}
\label{tab:node_clustering}
\resizebox{\textwidth}{!}{%
\begin{tabular}{|c|c|cccccccccccc|}
\hline
\multirow{2}{*}{Dataset} &
  \multirow{2}{*}{Metric} &
  \multicolumn{12}{c|}{Baselines} \\ \cline{3-14} 
 &
   &
  GCN &
  GAT &
  HGCN &
  HAN &
  MAGNN &
  GTN &
  HetSANN &
  HGT &
  GraphMSE &
  Simple-HGN &
  SHAN &
  \algname{} \\ \hline
\multirow{2}{*}{IMDB} &
  NMI &
  12.62$\pm$0.18 &
  13.12$\pm$0.34 &
  14.88$\pm$0.81 &
  15.54$\pm$0.84 &
  20.67$\pm$1.10 &
  19.98$\pm$0.38 &
  20.01$\pm$0.45 &
  19.66$\pm$0.34 &
  20.69$\pm$0.91 &
  22.38$\pm$0.74 &
  22.64$\pm$0.83 &
  \underline{\bf22.87$\pm$0.54}\\
 &
  ARI &
  12.80$\pm$0.29 &
  13.84$\pm$0.24 &
  15.71$\pm$0.78 &
  16.42$\pm$0.93 &
  21.60$\pm$0.48 &
  20.85$\pm$0.46 &
  21.75$\pm$0.51 &
  21.06$\pm$0.54 &
  21.47$\pm$0.80 &
  23.61$\pm$0.75 &
  26.49$\pm$0.46 &
  \underline{\bf26.83$\pm$0.63} \\ \hline
\multirow{2}{*}{DBLP} &
  NMI &
  76.20$\pm$0.12 &
  77.34$\pm$0.51 &
  78.18$\pm$0.96 &
  79.01$\pm$0.21 &
  80.39$\pm$0.37 &
  80.24$\pm$0.84 &
  80.87$\pm$0.35 &
  79.15$\pm$0.40 &
  36.32$\pm$0.46 &
  81.26$\pm$0.18 &
  \underline{\bf82.15$\pm$0.25} &
  81.34$\pm$0.31 \\
 &
  ARI &
  77.18$\pm$0.16 &
  78.36$\pm$0.63 &
  80.44$\pm$1.09 &
  82.51$\pm$0.33 &
  85.66$\pm$0.35 &
  83.39$\pm$0.76 &
  81.84$\pm$0.48 &
  82.41$\pm$0.52 &
  33.81$\pm$0.58 &
  85.73$\pm$0.21 &
  86.07$\pm$0.36 &
  \underline{\bf86.12$\pm$0.42} \\ \hline
\multirow{2}{*}{ACM} &
  NMI &
  55.52$\pm$0.37 &
  63.19$\pm$0.24 &
  64.93$\pm$0.50 &
  66.17$\pm$0.36 &
  70.86$\pm$0.44 &
  69.94$\pm$0.17 &
  69.10$\pm$0.20 &
  72.01$\pm$0.26 &
  72.82$\pm$0.39 &
  74.63$\pm$0.36 &
  75.19$\pm$0.27 &
  \underline{\bf75.38$\pm$0.19} \\
 &
  ARI &
  58.48$\pm$0.36 &
  65.63$\pm$0.32 &
  67.89$\pm$0.79 &
  68.92$\pm$0.33 &
  72.05$\pm$0.29 &
  71.46$\pm$0.24 &
  72.78$\pm$0.33 &
  77.09$\pm$0.36 &
  77.56$\pm$0.24 &
  78.68$\pm$0.48 &
  79.93$\pm$0.34 &
  \underline{\bf80.16$\pm$0.64} \\ \hline
\end{tabular}%
}
\vspace{-1em}
\end{table*}
%\vspace{-0.5em}
\subsection{Node Classification \& Clustering}
Node classification and clustering were performed ten times, and we report the average values and standard deviations of Macro-F1 and Micro-F1 for classification, as well as NMI and ARI for clustering. The classification and clustering were performed using SVM and $k$-means clustering on the embedding vectors of the labeled nodes.

As shown in Tables~\ref{tab:node_classification} and ~\ref{tab:node_clustering}, \algname{} performs better than the baseline models in most cases. The results from \algname{} and SHAN indicate the effectiveness of hyperbolic space embedding in heterogeneous graphs. Compared to SHAN, \algname{} effectively captures the continuous hierarchical structure of nodes contained within metapath instances that follow the metapath schema representing the semantic structure within heterogeneous graphs. Also, through effective aggregation of metapath-specific embeddings (Sec.~\ref{inter-metapath attention}), \algname{} captures important semantic information well. 

On the other hand, HGCN achieves better performance than GCN and GAT. The reason lies in that hyperbolic space has a constant negative curvature, and it expands exponentially. This property makes hyperbolic space more appropriate for modeling tree-like or hierarchically structured graphs where the number of nodes grows exponentially, as opposed to Euclidean space. At last, the results from heterogeneous GNNs show that these baselines outperform homogeneous GNNs by leveraging the semantic information derived from the heterogeneity of heterogeneous graphs.

\begin{figure}[t!]
\captionsetup[subfigure]{justification=centering}
\centering
        \begin{minipage}[b]{0.5\columnwidth}
                \centering
                \includegraphics[width=\linewidth]{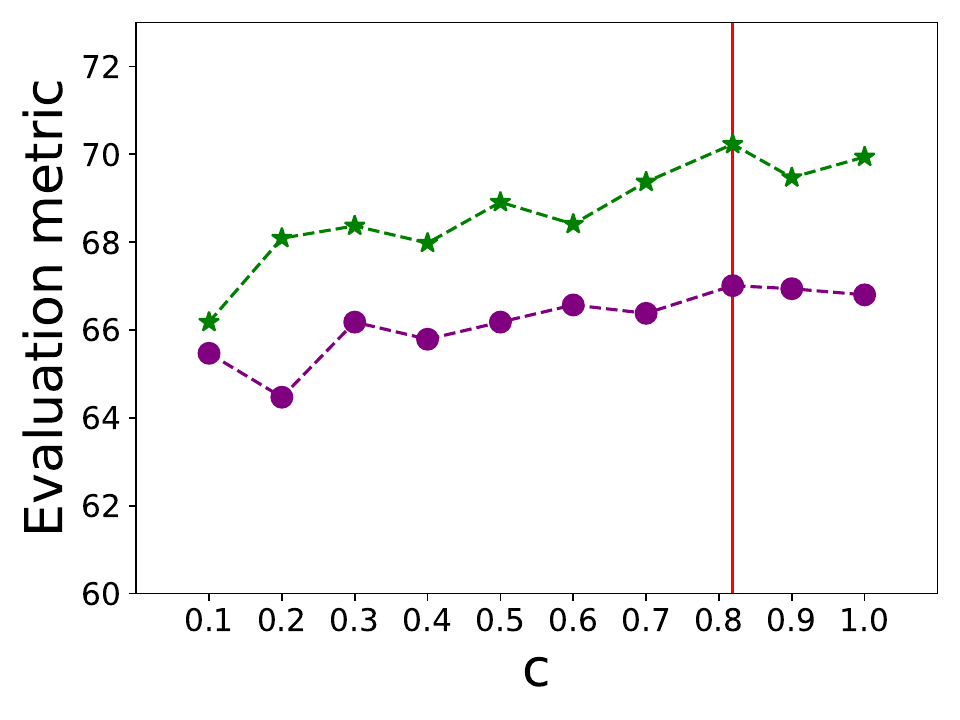}
                \vspace{-1.5\baselineskip}
                \subcaption{IMDB}
        \end{minipage}%
        \begin{minipage}[b]{0.5\columnwidth}
                \centering
                \includegraphics[width=\linewidth]{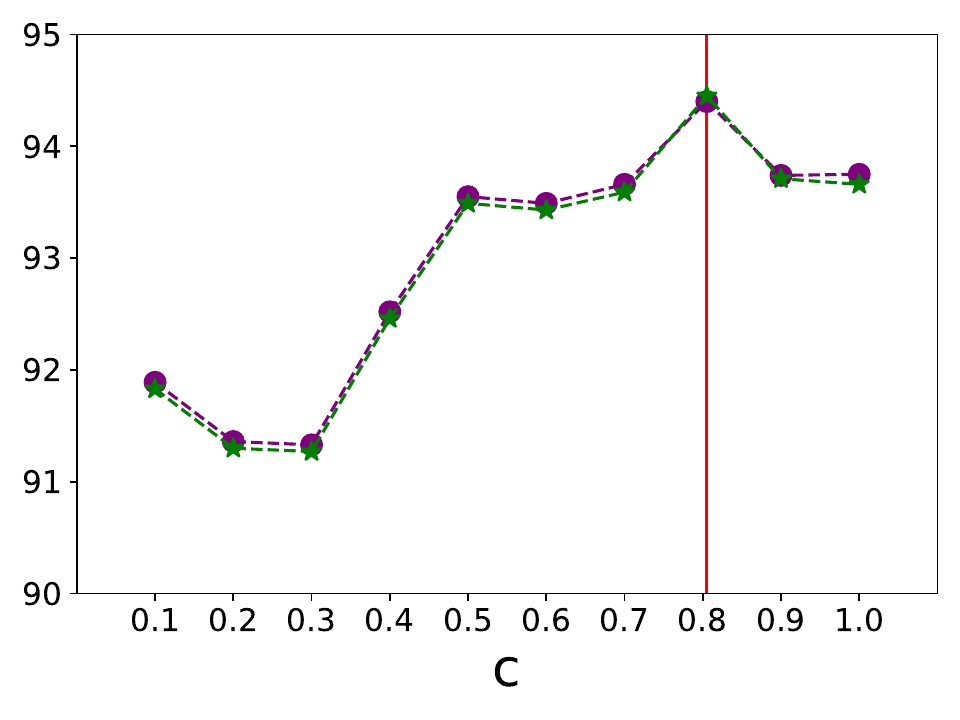}
                \vspace{-1.5\baselineskip}
                \subcaption{ACM}
        \end{minipage}
        \newline
        \begin{minipage}[b]{0.5\columnwidth}
                \centering
                \includegraphics[width=\linewidth]{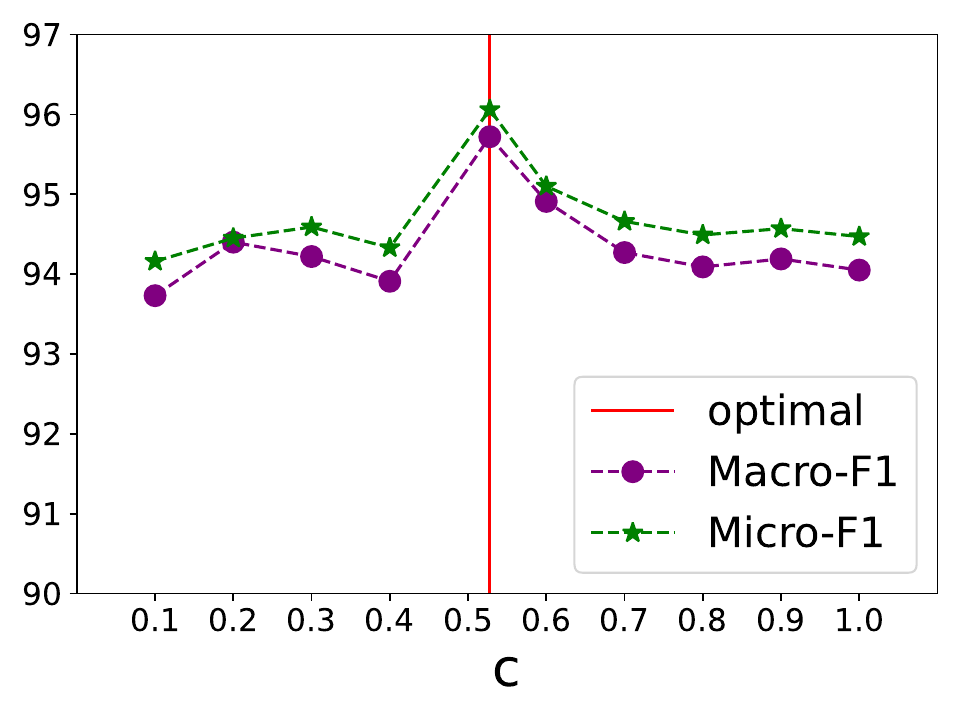}
                \vspace{-1.5\baselineskip}
                \subcaption{DBLP}
        \end{minipage}%
        \vspace{-1.5em}
        \caption{Node classification accuracy for varying the curvature parameter $c$. The red line indicates the optimal $c$.}
        \vspace{-1.5em}
        \label{fig:curvature_analysis}
\end{figure}
\vspace{-0.5em}
\subsection{Analysis of Curvature}
To analyze the effect of a curvature parameter $c$, that is the curvature is $-c$, we kept the curvature fixed during model training. %Note that the fixed curvature is not a learnable parameter.
As a result, Figure~\ref{fig:curvature_analysis} illustrates the node classification accuracy for different curvatures. The red line highlights the optimal curvatures for each dataset, which are $-0.8189, -0.5278,$ and $-0.8052$ for IMDB, DBLP, and ACM, respectively.

Figure~\ref{fig:curvature_analysis} shows that \algname{} achieved the highest performance with the optimal curvature. As $c$ approaches zero, we observed that the performance of \algname{} generally decreases. This is due to the hyperbolic space becoming similar to the flat curvature of the Euclidean space, which weakens its property to represent hierarchical structures. On the other hand, when the curvature parameter $c$ becomes larger than the optimal value and continues to increase, the hyperbolic space with that curvature does not appropriate well with the power-law distribution inherent in each dataset. This mismatch leads to lower performance compared to the optimal case. From this empirical observation, we conclude that setting the curvature as a learnable parameter and training it to align with the power-law distribution inherent in each dataset is an effective approach.
\vspace{-0.5em}
\section{Conclusion}
In this paper, we proposed the hyperbolic heterogeneous graph attention networks. Without domain knowledge, \algname{} samples metapath instances and leverages them in hyperbolic spaces for effective heterogeneous graph representation learning. We conduct extensive experiments using three real-world heterogeneous graph datasets. The results demonstrate that \algname{} performs better than other state-of-the-art baselines. For the future work, we plan to explore optimizing curvature with regularization to enhance the representation of heterogeneous graphs, aiming to better capture the inherent power-law distribution.
\vspace{-0.5em}
\iffalse
\begin{acks}
This work was supported by the National Research Foundation of Korea (NRF) grant funded by the Korea government (MSIT) (No. RS-2023-00214065) and by the Institute of Information \& Communications Technology Planning \& Evaluation (IITP) grant funded by the Korea government (MSIT) (No. RS-2022-00155857, Artificial Intelligence Convergence Innovation Human Resources Development (Chungnam National University)).% and Basic Science Research Program through the National Research Foundation of Korea (NRF) funded by the Ministry of Education (RS-2023-00272241).
\end{acks}
\fi
%\vspace{-0.5em}
\bibliographystyle{ACM-Reference-Format}
\balance
\bibliography{WWW2024_Submission.bbl}

\end{document}